\documentclass[10pt,twocolumn,letterpaper]{article}

\usepackage{iccv}
\usepackage{times}
\usepackage{epsfig}
\usepackage{graphicx}
\usepackage{amsmath}
\usepackage{amssymb}
\usepackage[binary-units=true]{siunitx}
\usepackage{float}
\usepackage{stfloats}
\usepackage{placeins}
\usepackage[pagebackref=true,breaklinks=true,letterpaper=true,colorlinks,bookmarks=false]{hyperref}
\iccvfinalcopy 

\def\iccvPaperID{} 

\ificcvfinal\fi
\begin{document}

\title{\vspace{-2ex}Image-Based Parking Space Occupancy Classification: Dataset and Baseline}

\author{Martin Marek\\
ParkDots, PosAm\\
{\tt\small martin.marek1999@gmail.com}
}

\maketitle

\begin{abstract}
We introduce a new dataset for image-based parking space occupancy classification: {\normalfont ACPDS}. Unlike in prior datasets, each image is taken from a unique view, systematically annotated, and the parking lots in the train, validation, and test sets are unique. We use this dataset to propose a simple baseline model for parking space occupancy classification, which achieves 98\% accuracy on unseen parking lots, significantly outperforming existing models. We share our dataset, code, and trained models under the MIT license: \url{https://github.com/martin-marek/parking-space-occupancy}.
\end{abstract}

\section{Introduction}

Live information about parking space occupancy can be used to effectively navigate drivers, reducing congestion, emissions, and saving time.

There are two common approaches to monitor the occupancy of individual parking spaces: installing a sensor on every parking space or using a camera to monitor multiple parking spaces at once. In general, if a single camera can capture tens of parking spaces, the cost per parking space can be significantly lower compared to a parking sensor.

A camera can be used in two basic ways to monitor the occupancy of parking spaces. One way is to capture and process a live video feed allowing for methods such as motion tracking \cite{ke, cai_deep_2019, slot_det}. The other way is to capture images at a larger interval and process them individually \cite{pklot, cnrpark, barry_street, drone}. In this study, we focus only on image-based models, for several reasons. First, an image can be captured using a long exposure time, allowing for good color reproduction even in low-light; in contrast, capturing a video limits the camera's exposure time to the inverse of the framerate, likely 1/10 of a second or shorter. Additionally, a burst of images can be merged to increase dynamic range, especially in direct sunlight \cite{hdr_plus}. Second, if inference is done off-camera, capturing images at longer intervals significantly decreases data flow. Third, if the camera feed temporarily failed, an image-based model could immediately recover, as it does not rely on history; in contrast, a video-based model might lack the necessary context for inference. Lastly, it is easier to capture and annotate a dataset composed of images, compared to videos.

In prior works on camera-based parking space occupancy classification, the authors generally train their model on a large but generic object-detection dataset \cite{ke, cai_deep_2019, slot_det} or on an application-specific dataset consisting of just a few parking lots \cite{pklot, cnrpark, barry_street}. To evaluate model generalization, the authors use a separate dataset consisting of previously-unseen parking lots, achieving 89\% - 96\% accuracy.

We introduce a challenging new dataset where each image is taken from a unique view, corresponding to over 11,000 unique parking space annotations -- almost an order-of-magnitude more than the largest previously-published dataset \cite{cnrpark}. Our dataset is the first publicly available dataset that directly tests model generalization, by using separate parking lots for the train, validation, and test sets. It is also the first dataset with a consistent annotation format, enabling new augmentations and pooling methods. We use this dataset to develop a simple baseline model that achieves over 98\% accuracy on previously-unseen parking lots, significantly outperforming existing models.

\begin{figure}[!t]
    \centering
    \includegraphics[width=1.0\linewidth]{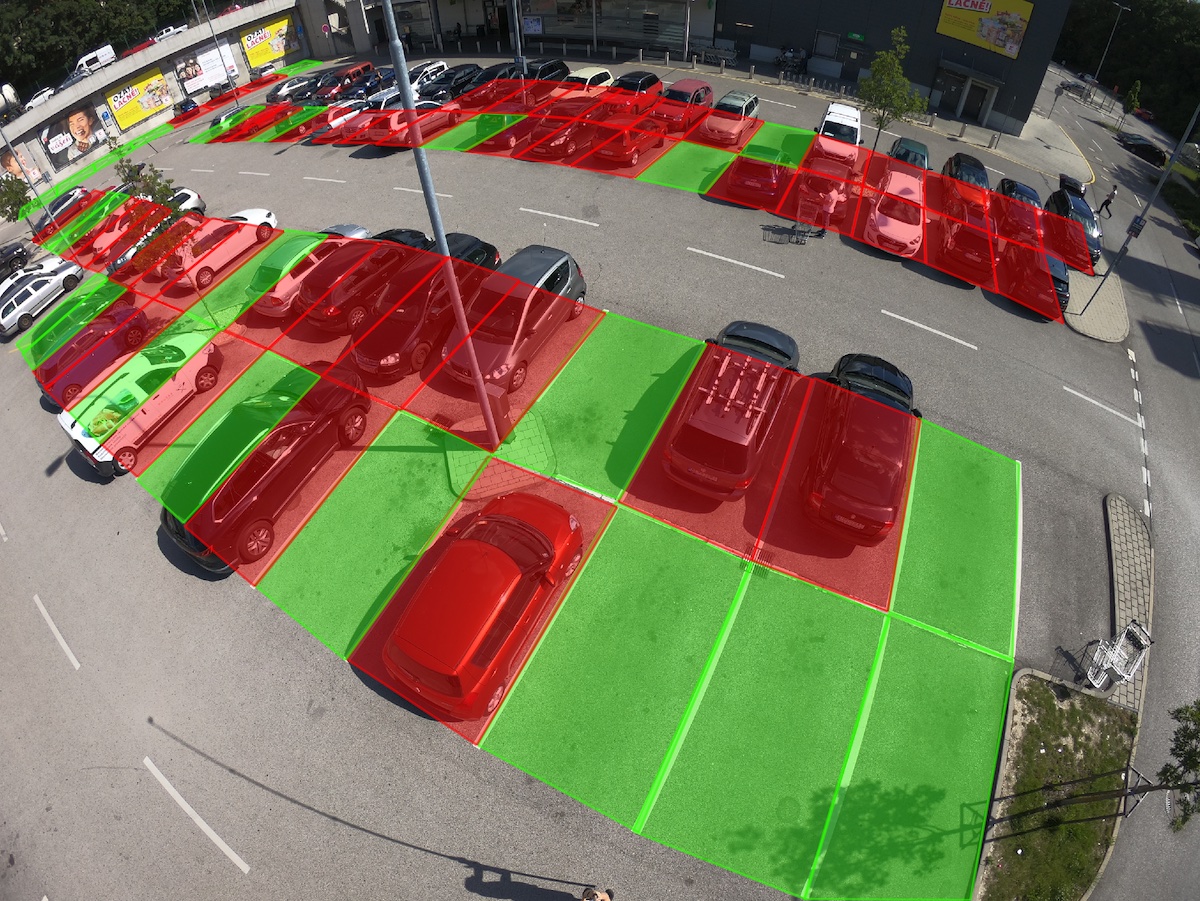}
    \caption{A sample image from our dataset. Each parking space is annotated using a quadrilateral corresponding to the edges of the parking space.}
    \label{fig:ds_sample}
\end{figure}

\section{Related work}

Existing models can be split into two main groups: video-based and image-based.

\subsection{Video-based models}

Video-based models rely on a continuous video feed to perform inference.

Ke \etal \cite{ke} use a Single Shot MultiBox Detector (SSD) \cite{ssd} together with a standardized tracking algorithm \cite{sort} and match the detected objects to parking spaces. Their model achieves 95.6\% accuracy in a parking garage not included in the training set, but the authors have adjusted model parameters to optimize model performance for this garage.

Cai \etal \cite{cai_deep_2019} use Mask-RCNN \cite{mask_rcnn} combined with a memory characterizing features of past cars detected. They build their own dataset for evaluation and achieve 88\% sensitivity (the authors do not state the accuracy of their model).

Li \etal \cite{slot_det} take a fundamentally different approach to locating vacant parking spaces. They utilize a surround camera system mounted on a car to detect vacant parking spaces. They build their own model and dataset and achieve precision and recall of 98\% and 92\%, respectively (the authors do not state the accuracy of their model).

\subsection{Image-based models}

Almeida and Oliveira \cite{pklot} introduce a dataset named \textit{PKLOT}, consisting of 12,417 images, taken from 3 views of 2 parking lots. However, they suggest that the dataset be split into train and test sets based on the date of capture of each image, \textit{not} based on the parking lot or parking space. As a result, both training and evaluating a model on this dataset will lead to an overestimated accuracy.

Amato \etal \cite{cnrpark} introduce a new dataset of roughly 150,000 images captured from 9 views of a single parking lot containing 164 parking spaces. They train a classifier on this dataset and achieve 98\% accuracy when evaluated on the same dataset. However, to test the generalization performance of their model, they also evaluate it on the \textit{PKLOT} dataset, but only achieve 93.7\% accuracy.

Acharya \etal \cite{barry_street} train a classifier on \textit{PKLOT} and evaluate it on their own dataset, achieving 96.6\% accuracy. The reason they achieve a relatively high accuracy is likely because their evaluation dataset only contains 30 parking spaces and no occlusions.

Hsieh \etal \cite{drone} take a different approach to monitor parking lot occupancy: they use a drone to detect and count vehicles. They build their own dataset of drone images, capturing 4 parking lots and nearly 90,000 cars, and annotate these images using bounding boxes. They use an object detector with a custom region proposal network to detect vehicles in images. The final output of their model is vehicle locations and the total count of vehicles; they do not classify the occupancy of individual parking spaces. Their model scores a mean absolute error of 22 for the vehicle count.

\section{Dataset}

\begin{figure*}
    \centering
    \includegraphics[width=1\linewidth]{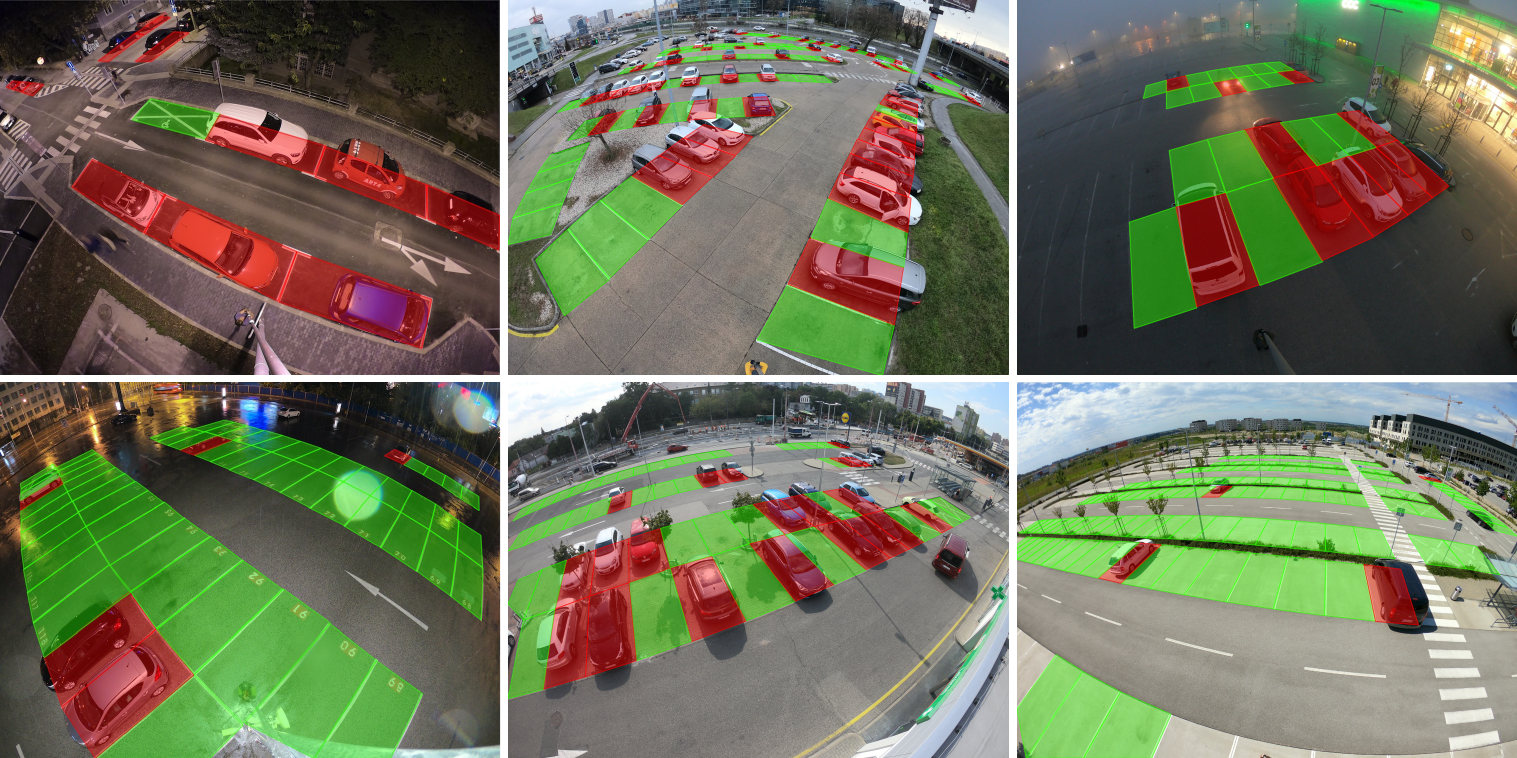}
    \caption{A sample of annotated images from our dataset. We collected images from different parking lots, under varied weather and lighting conditions, at different occupancy levels.}
    \label{fig:dataset}
\end{figure*}

We introduce a new dataset for parking space occupancy classification: \textit{Action-Camera Parking Dataset (ACPDS)}. The goal of our dataset is to improve and correctly test for model performance on previously unseen parking lots. To this end, the dataset captures various parking lots, each image is captured from a unique view, systematically annotated, and unique parking lots are used for the train, validation, and test sets. As a result, the validation and test accuracy obtained on our dataset correspond to model performance on a previously unseen parking lot.

\subsection{Data collection}

We mounted a GoPro Hero 6 action camera to a 12-meter telescoping pole and used a smartphone to view a live feed from the camera and control the shutter. This enabled us to walk around with the setup and capture each image from a unique view. We captured tens of different parking lots and streets, under various weather and lighting conditions (see Figure \ref{fig:dataset}). However, we notably didn't capture any images that include snow.

Each image is captured by the same camera in the ``wide'' field of view setting, in full (4000 x 3000) resolution. Moreover, each image is captured from a roughly 12-meter height, corresponding to a common height of lamp posts. We consider this similarity to be crucial for practical applications. If a camera were installed on a lamppost, it could typically be at most 12 meters above the ground, resulting in the same view angles and same levels of occlusions as what we have captured. In contrast, a dataset captured from high above the ground would not include significant occlusions, and any model trained on such a dataset would generalize poorly to a low installation height (with strong occlusions between individual parking spaces).

\subsection{Annotation}

We labeled each image in the same way, as illustrated in Figure \ref{fig:ds_sample}, using Labelbox. To annotate one parking space, we drew a quadrilateral around it, ensuring that the edges of the parking space and the quadrilateral are aligned. In a few instances, however, when part of a parking space was cut off at the edge of an image, the shape of the visible parking space projection became a pentagon. For consistency, even in these instances, we labeled such parking spaces using a quadrilateral. This resulted in 2 of the 4 edges of the annotated quadrilateral not being aligned with the parking space. This is visible on the bottom side of each image in Figure \ref{fig:pred}.

We only labeled those parking spaces where we were confident about their coordinates and occupancy. Often, we needed to rely on the consistency of the parking lot layout: even when a parking space was fully occluded, we were still able to label its coordinates by knowing the coordinates of the surrounding parking spaces. Other times, when vehicles were heavily occluded, we needed to count them and compare their relative positions: only this allowed us to decide on the occupancy of a given parking space. Both of these challenges can be seen in Figure \ref{fig:ds_sample}, especially on the left side of the image, where heavy occlusions are present.

We checked each annotation in the train set at least once; in the validation and test sets, we checked each annotation at least twice.

\subsection{Dataset size}

We captured 293 images containing 11,236 unique views of a parking space, spread across tens of different parking lots and streets. 5,376 of these parking spaces were occupied, corresponding to 48\% of all parking spaces. We used 231, 35, and 27 images for the train, validation, and test sets, respectively. While this may sound like a low number of images, both the validation and test sets contain over 1,400 unique views of a parking space. The relatively high costs of capturing and annotating images using our methodology prevented us from building a larger dataset.

\section{Models}

\begin{figure*}
    \centering
    \includegraphics[width=1.0\linewidth]{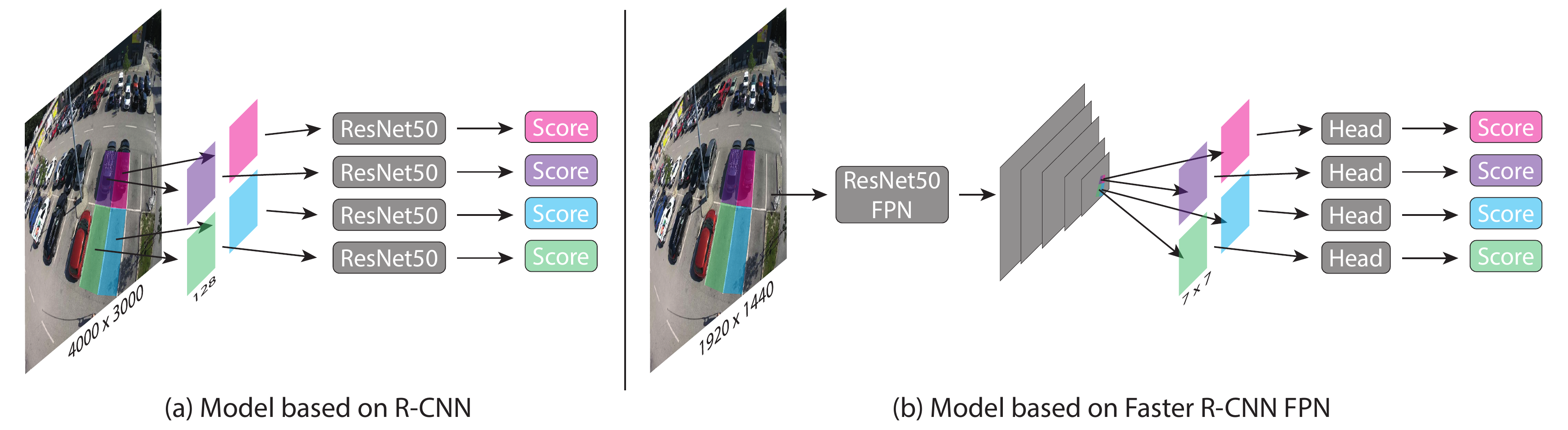}
    \caption{We propose two models for parking space occupancy classification, both inspired by two-stage object detectors. We replace region proposals by the coordinates of parking spaces. The output of our models is an occupancy score for each parking space.}
    \label{fig:arch}
\end{figure*}

In addition to building a new dataset, we design and train two simple models on this dataset. We intend these models to function as a simple baseline for our dataset.

Our models are inspired by object detectors. However, regular object detectors cannot be applied directly to our dataset to classify parking space occupancy. Both single-stage and two-stage object detectors rely on non-maximum suppression with an arbitrary intersection-over-union (IoU) threshold to filter proposals/detections. There is no clear IoU threshold for our application: for unoccluded parking spaces, we would prefer a high IoU threshold such that we don't double-count a single vehicle; on the other hand, to detect heavily occluded vehicles, the IoU threshold must be very low. Single-stage detectors are further constrained by using a set of default bounding boxes. Even if we solved the aforementioned limitations (which has been achieved by Carion \etal \cite{detr}, for example), we would need to assign each detected vehicle to a specific parking space. This is not a trivial issue: reliably assigning occluded vehicles to specific parking spaces would require knowledge of the camera calibration. Lastly, using a regular object detector would be a missed opportunity to take into account our knowledge of the parking lot layout -- before passing the parking lot image through our model, we already know which regions we are interested in, based on the location of each parking space.

To leverage our knowledge of parking space locations, we implement two custom models inspired by two two-stage object detectors (R-CNN \cite{rcnn}, and Faster R-CNN FPN \cite{frcnn, fpn}). In both of these models, we use the annotated parking space coordinates as the regional proposals. We discuss the details below.

\subsection{R-CNN}

Our first proposed model takes inspiration from R-CNN \cite{rcnn}, as well as prior models used for image-based parking space occupancy classification \cite{pklot, cnrpark, barry_street}. First, we pool image patches corresponding to each parking space directly from the image. Afterward, these patches are passed separately through a binary classifier (ResNet50 \cite{resnet}). The output of the classifier is the occupancy score for each parking space. The model is illustrated in part (a) of Figure \ref{fig:arch}. We discuss our pooling technique in section \ref{sec:pooling}.

While models based on the R-CNN architecture are no longer considered efficient for general-purpose object detection, our specific application actually makes the architecture desirable. First, we capture images in (4000 x 3000) resolution. An image of this resolution is too large to be passed directly through a deep convolutional network, but we can easily pass small image patches through a ResNet50. In the original R-CNN model, the number of region proposals is large (2,000), and the pooling resolution of each patch is high (224 x 224). In our application, however, we pass at most around 100 image patches through the classifier, and as we show in section \ref{sec:results}, we can get away with a patch resolution of just (128 x 128). These differences combine to a roughly 60-fold inference time decrease compared to the original R-CNN model. We further discuss inference time in section \ref{sec:speed}.

A notable disadvantage of this model is that there can never be any information flow between pooled image patches. This might make it very difficult for the model to reason about occlusions.

\subsection{Faster R-CNN FPN}

Our second proposed model takes inspiration from Faster R-CNN FPN \cite{frcnn, fpn}. First, we pass a resized image through a ResNet50 combined with a feature pyramid network \cite{fpn}. Afterward, we pool features corresponding to each parking space from the feature pyramid and pass them separately through a classification head to obtain the final occupancy scores. The model is illustrated in part (b) of Figure \ref{fig:arch}. We use the same heuristic as Lin \etal x\cite{fpn} to decide which pyramid layers to pool from. We also use the same pooling resolution (7 x 7).

Unlike our first proposed model, this model cannot utilize the full resolution of our dataset. On the other hand, the architecture does allow for information flow between parking spaces, so we believe there is a better potential for the model to reason about occlusions.

\subsection{Pooling} \label{sec:pooling}

Both of our proposed models rely on pooling features corresponding to each parking space. Here, we discuss two different methods to perform the pooling.

\begin{figure}[H]
    \centering
    \includegraphics[width=1.0\linewidth]{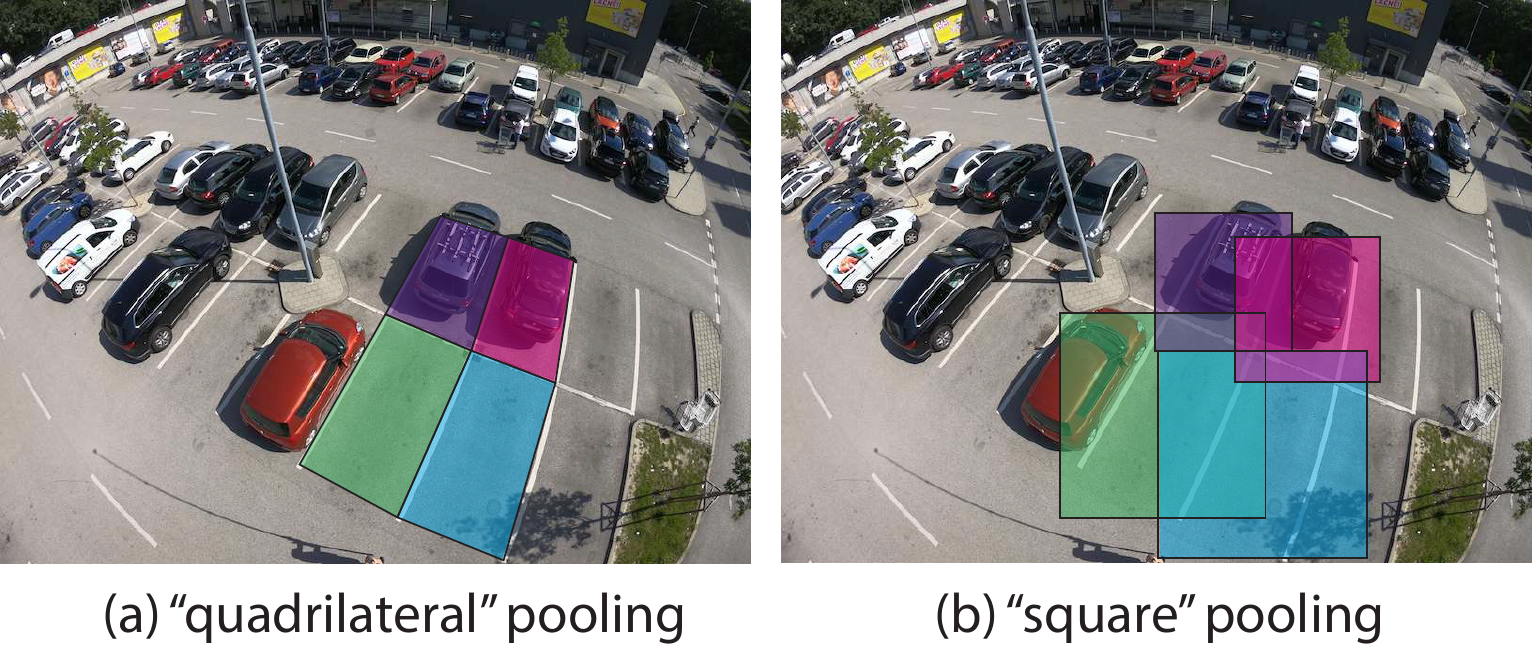}
    \caption{We illustrate our two proposed methods to pool features from parking spaces. The parking space annotations are drawn in image (a), using 4 colored quadrilaterals. In method (a), we interpolate pixels from these quadrilaterals directly. In method (b), we consider a minimum bounding square around each parking space and interpolate pixels from these squares.}
    \label{fig:pooling}
\end{figure}

Since we annotate each parking space using a quadrilateral, our first proposed pooling method is to warp the pixels inside each quadrilateral and project them to an (S x S) patch directly. Our second proposed method is to consider a minimum bounding square around each parking space annotation and interpolate this square to an (S x S) patch. We illustrate the two methods in Figure \ref{fig:pooling}.

A notable difference between these pooling methods is that one only pools information from within a parking space quadrilateral, while the other also pools information from surrounding pixels. When the quadrilateral has an uneven aspect ratio (\eg, near an image edge) or it includes heavy occlusions, the context obtained from the surrounding pixels can be useful. This difference is especially relevant for the model based on R-CNN; the model based on Faster R-CNN FPN can utilize its backbone to obtain context from surrounding pixels either way.

We implement both of these pooling methods natively in PyTorch \cite{pytorch} with CUDA and TorchScript support. This allows for inference on GPU, CPU, and mobile.

\section{Evaluation}

We train both of our models with different hyperparameter configurations on our dataset. We do not train our models on any other datasets. To the best of our knowledge, our dataset is currently the only publicly available dataset with precise annotations for each parking space edge. Other datasets typically draw squares \cite{cnrpark} or quadrilaterals \cite{pklot} around each parking space without any consistency; our models are not intended for this label format. Moreover, prior works used a separate dataset for training and evaluation to test model generalization. Since our dataset is already split into train, validation, and test sets by parking lots, we do not need a second dataset to test for generalization.

\subsection{Training details}

For consistency, both of our models use a pre-trained ResNet50 backbone. The model based on R-CNN uses weights from ImageNet \cite{imagenet} training, while the model based on Faster R-CNN FPN uses weights obtained on the COCO dataset \cite{coco}. The reason behind this difference is that the model based on R-CNN is essentially just a regular classifier, apt for Imagenet training, whereas training a model with a feature pyramid network requires the use of an object detection dataset (COCO). We obtained these weights from the torchvision package.

Following a widely adopted practice in object detection, our backbone batch normalization weights and statistics are frozen. We additionally freeze layers 1 and 2 in our backbone, to reduce overfitting to our small dataset.

We train both of our models in each configuration using AdamW \cite{adamw}, with a learning rate of $10^{-4}$ for 50 epochs, and a learning rate of $10^{-5}$ for additional 50 epochs. We are aware that SGD with momentum is the standard optimizer for vision tasks and typically results in better generalization than adaptive methods \cite{optimizer_generalization}. We did not perform an exhaustive test of different optimizers and their respective parameters. However, from a few simple training runs, we found AdamW to result in better validation loss than SGD with momentum. It is plausible that this difference would disappear if we performed a more exhaustive search for optimizer parameters.

We find that training our models with a high learning rate for the first 50 epochs and then dropping the learning rate results in better validation loss as compared to training with a constant learning rate throughout. This is consistent with published results \cite{large_lr}. Since we use relatively high-resolution images for training, we only use a mini-batch of size 1. This allows us to perform training on a GPU with \SI{11}{\giga\byte} memory. 

\begin{figure}[H]
    \centering
    \includegraphics[width=1.0\linewidth]{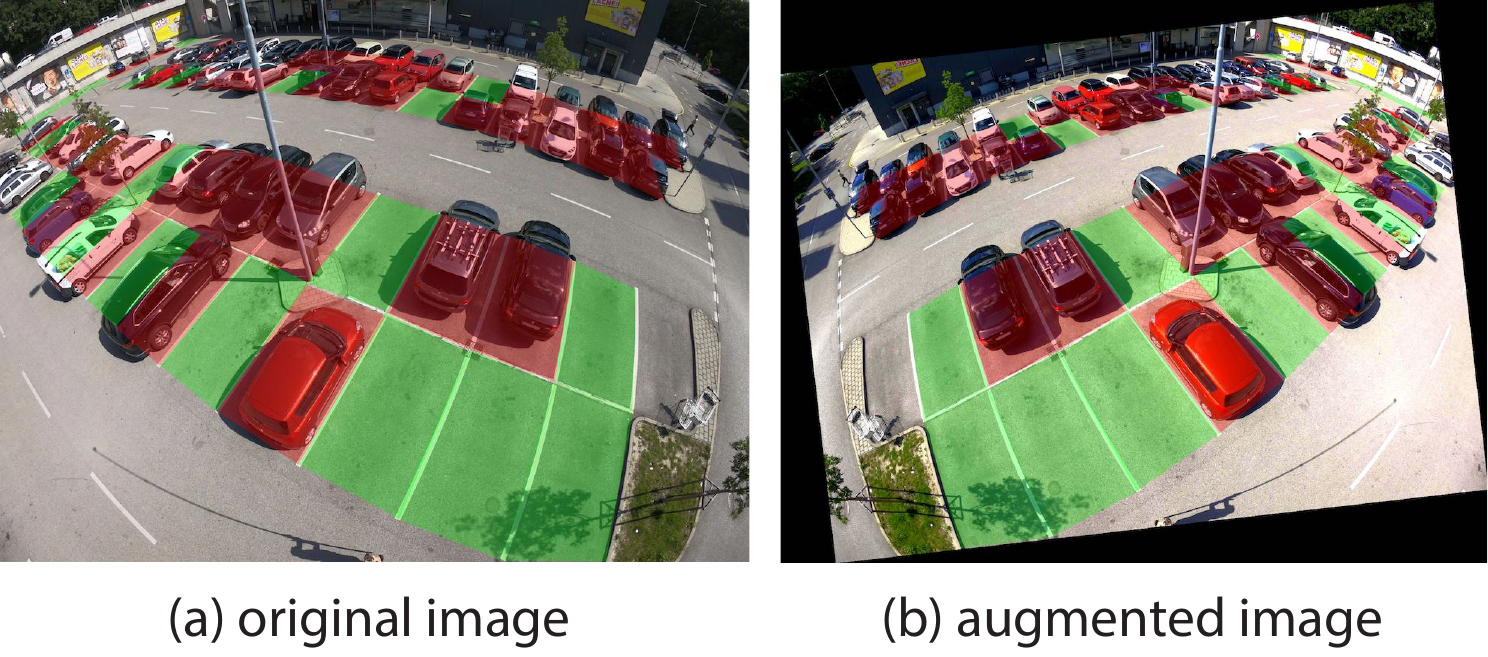}
    \caption{We augment each training example using a random left-right flip, random rotation, and a random adjustment to brightness, contrast, saturation, and hue.}
    \label{fig:augmentation}
\end{figure}

We augment each training example using a random left-right flip, random rotation, and a random adjustment to brightness, contrast, saturation, and hue. We show an example of the augmentation in Figure \ref{fig:augmentation}. In order to rotate the image, it is important that our parking space annotations are in the form of precise quadrilaterals. This way, we can find the correct minimum bounding square for each parking space (as described in section \ref{sec:pooling}) even after rotating our annotations. In contrast, if we labeled parking spaces using minimum bounding squares in the first place, and we rotated the image together with the annotations, these squares would no longer represent the minimum bounding squares for each parking space. The exact parameters of all the mentioned augmentations can be found in our code.

In order to achieve fast training speeds, we cache the whole dataset in main memory, without any augmentations, and perform all augmentations on the GPU, before passing a mini-batch through the model. These optimizations enable us to train for a full epoch with (4000 x 3000) images in around 40 seconds, on an Nvidia RTX 2080 Ti GPU.

\begin{table*}[!bp]
\begin{center}
\begin{tabular}{|c|c|c|c|c|}
\hline
     Architecture &        Pooling &  Resolution & Valid. accuracy [\%] & Test accuracy [\%] \\
\hline\hline
 \textbf{Faster R-CNN FPN} & \textbf{square} & \textbf{1440} & \textbf{98.58 $\pm$ 0.07} & \textbf{98.51 $\pm$ 0.10} \\
 Faster R-CNN FPN &         square &        1100 &     98.54 $\pm$ 0.11 &   98.52 $\pm$ 0.11 \\
 Faster R-CNN FPN &         square &         800 &     98.36 $\pm$ 0.07 &   98.31 $\pm$ 0.08 \\
 Faster R-CNN FPN &  quadrilateral &        1440 &     98.34 $\pm$ 0.10 &   98.31 $\pm$ 0.09 \\
 Faster R-CNN FPN &  quadrilateral &        1100 &     98.28 $\pm$ 0.14 &   98.00 $\pm$ 0.08 \\
 Faster R-CNN FPN &  quadrilateral &         800 &     97.80 $\pm$ 0.07 &   97.97 $\pm$ 0.14 \\
            R-CNN &         square &         256 &     98.11 $\pm$ 0.07 &   97.62 $\pm$ 0.07 \\
\textbf{R-CNN} & \textbf{square} & \textbf{128} & \textbf{98.38 $\pm$ 0.06} & \textbf{97.97 $\pm$ 0.07} \\
            R-CNN &         square &          64 &     98.00 $\pm$ 0.08 &   97.73 $\pm$ 0.13 \\
            R-CNN &  quadrilateral &         256 &     96.39 $\pm$ 0.15 &   96.08 $\pm$ 0.12 \\
            R-CNN &  quadrilateral &         128 &     96.27 $\pm$ 0.17 &   96.63 $\pm$ 0.15 \\
            R-CNN &  quadrilateral &          64 &     95.87 $\pm$ 0.11 &   96.39 $\pm$ 0.17 \\
\hline
\end{tabular}
\end{center}
\caption{Accuracy for different model configurations. The resolution column refers in case of R-CNN to the pooling resolution, and in case of Faster R-CNN FPN to the input image resolution. The uncertainty estimate is an estimated standard error from 5 training runs.}
\label{tab:acc}
\end{table*}

\subsection{Results} \label{sec:results}

We have trained both of our models in each configuration 5 times, each time with random initialization. We use these 5 training runs to report the uncertainty in each of our experiments, using the mean accuracy and its estimated standard error.

For both models, we test both pooling methods from section \ref{sec:pooling}. For the model based on R-CNN, we test 3 pooling resolutions: \{64, 128, 256\}. For the model based on Faster R-CNN FPN, we test 3 input resolutions, defined by the size of the smaller image edge: \{800, 1100, 1440\}. We do not alter any other hyperparameters, since our models are heavily based on existing, well-researched architectures.

We report the validation and test accuracy for each model configuration in Table \ref{tab:acc}. The configuration with the highest validation accuracy is highlighted for both architectures. We consider it important to select a model configuration based on the validation set accuracy and leave the test accuracy as an unbiased estimate of model generalization.

We observe that in general, the Faster R-CNN FPN architecture performs better than the R-CNN architecture. We find square pooling to be always preferable to quadrilateral pooling, especially for the R-CNN architecture. For the R-CNN architecture, we find the highest accuracy at resolution (128 x 128). This is also the resolution we expected to perform the best: the smallest parking spaces in our dataset take up around (100 x 100) pixels and we want to minimize any patch upsampling; at the same time, we want to avoid discarding useful information by choosing too small a resolution (\eg, 64 x 64). For the Faster R-CNN FPN architecture, we found the highest resolution (1920 x 1440) to perform the best. This is still a significantly lower resolution than what our camera captured (4000 x 3000), but our GPU memory prevented us from testing a higher resolution. Moreover, using a large input resolution significantly increases inference time -- we discuss this in section \ref{sec:speed}.

It is also interesting to observe that the Faster R-CNN FPN architecture generalizes better to the test dataset than the R-CNN architecture. We suspect this is because the test set systematically differs from the validation set (\eg, in the number and strength of occlusions).

\begin{figure}[H]
    \centering
    \includegraphics[width=1\linewidth]{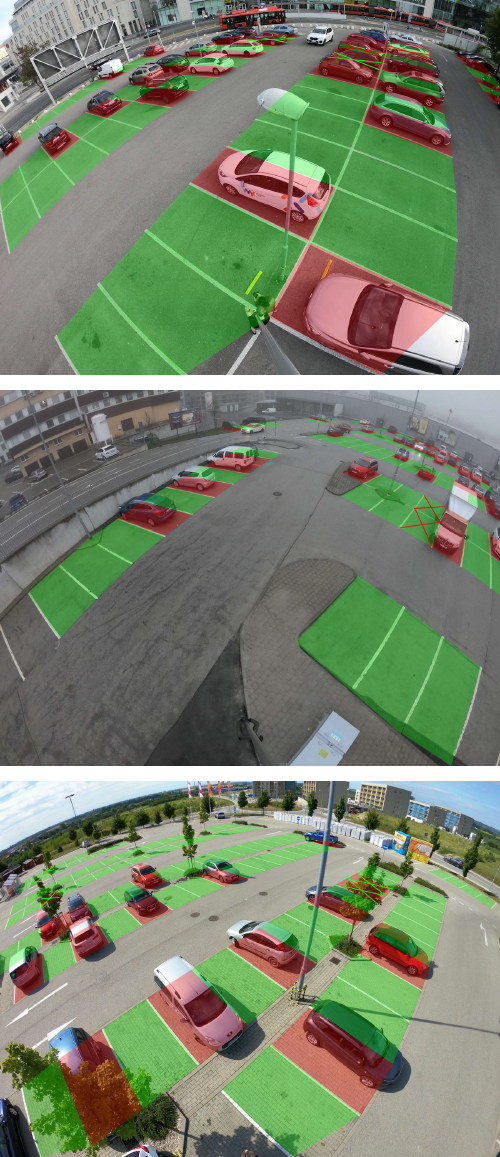}
    \caption{Predictions by our model based on Faster R-CNN FPN for 3 challenging images. Model predictions are shown by transparent quadrilaterals; where these predictions differ from the labels, the labeled occupancy is displayed by a large cross across the whole parking space. The top image shows heavy occlusions by cars; the middle image shows a large vehicle taking up multiple parking spaces; the bottom image shows an occlusion by trees.}
    \label{fig:pred}
\end{figure}

We show predictions for the model with the highest validation accuracy for 3 challenging images in Figure \ref{fig:pred}. Each of the images in Figure \ref{fig:pred} represents a failure of the model. In the top image, the model fails to reason about heavy occlusions caused by surrounding cars. It is not clear to us whether this problem would go away if we built a larger dataset or whether this would also require a new model architecture. On the other hand, the failure in the bottom image is simply caused by our dataset not including enough parking spaces occluded by trees; we are confident we could fix this by building a larger dataset or implementing a more efficient training method (\eg, using generated occlusions as a form of data augmentation). The middle image shows a large vehicle taking up multiple parking spaces; this is very rare in our dataset.

\begin{figure}[H]
    \centering
    \includegraphics[width=0.95\linewidth]{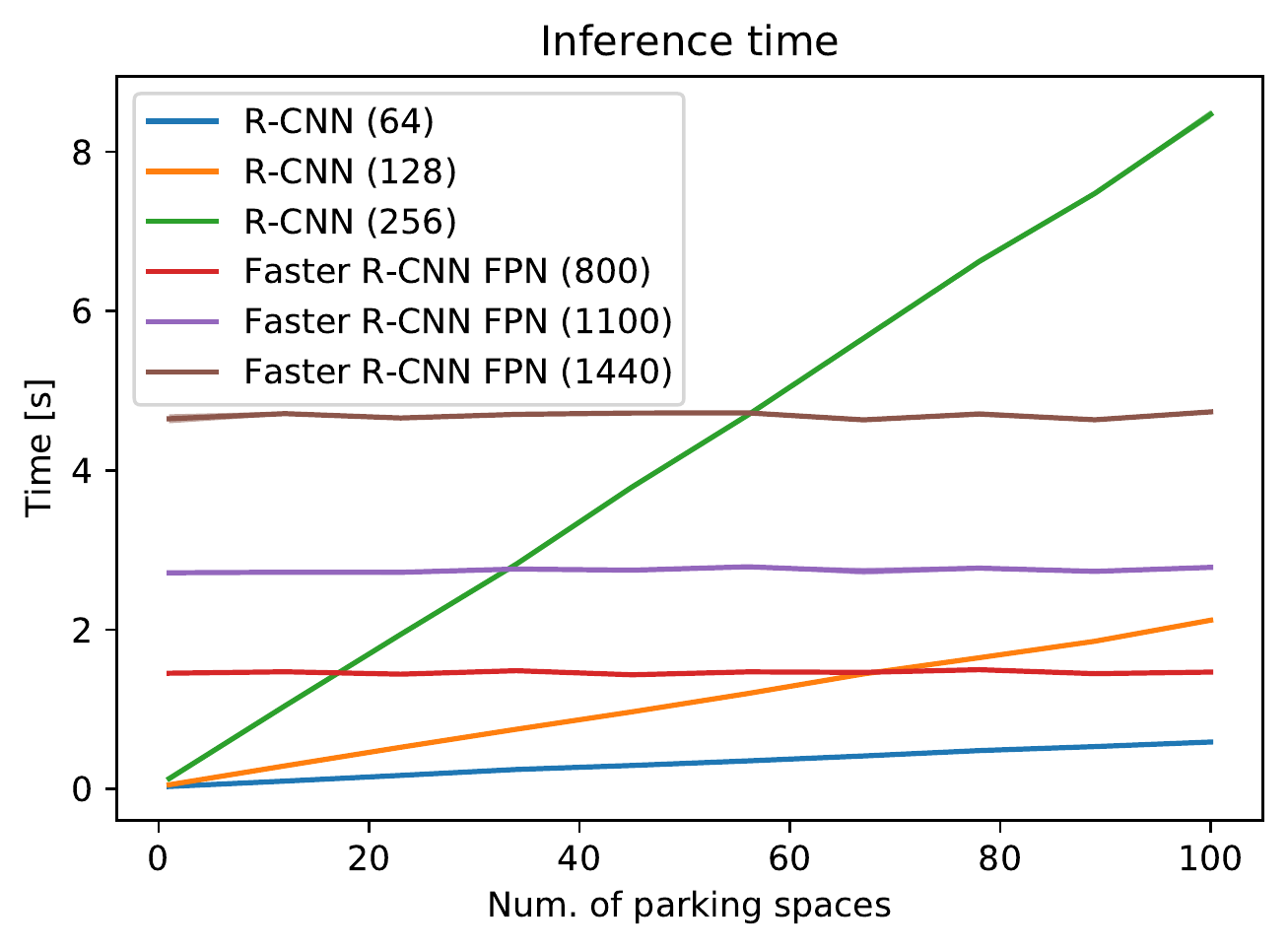}
    \caption{Inference time for various model configurations on an Intel Core i9-9900K CPU.}
    \label{fig:speed}
\end{figure}

\subsection{Inference time}\label{sec:speed}

In Figure \ref{fig:speed}, we compare the inference time of our proposed models under different configurations, as we vary the number of parking spaces. We measure the inference time on an Intel Core i9-9900K CPU, to simulate CPU deployment. We observe that models based on the R-CNN architecture have an almost perfectly linear relationship between the number of parking spaces and the total inference time. In contrast, for models based on the Faster R-CNN FPN architecture, passing an image through the backbone is a very expensive operation, while pooling and passing features through a classification head takes negligible compute. As a result, these models have an almost perfectly constant inference time.

\subsection{Model comparison}

We consider the R-CNN architecture preferable for practical deployment. While it does not perform as well as the Faster R-CNN FPN architecture on the test set, our dataset is intentionally very challenging; for unoccluded parking spaces, we would expect both of our models to achieve accuracy over 99\%. For real-world applications, the R-CNN architecture is the more flexible one. It should be possible to use a very high-resolution camera to capture hundreds of parking spaces at once; the model only cares that each pooled image patch is of high-enough resolution. Conversely, it should also be possible to use a low-resolution camera to capture only a few parking spaces, as long as the resolution of each pooled image patch is high enough. In contrast, the Faster R-CNN architecture requires a specific input resolution. It might generalize poorly to new resolutions and it is constrained by memory and compute at very high resolutions.

\section{Conclusion}

We present a new dataset for parking space occupancy classification, \textit{ACPDS}, where each image is taken from a unique view. Our dataset is split into train, validation, and test sets based on parking lots and has a consistent annotation format. As a result, our dataset allows for good generalization \textit{and} tests generalization directly. We design and train a practical model on this dataset and achieve accuracy over 98\%, significantly outperforming existing models. Our model is suitable for CPU deployment and can be used with images of different resolutions. The model has learned to tolerate moderate occlusions but fails under heavy occlusions. We intentionally made our dataset challenging by including heavily occluded parking spaces, so that future models can continue to be benchmarked on this dataset, and improve upon our results. Thanks to our streamlined data collection and annotation process, our dataset can also be extended by other researchers.

\FloatBarrier
{\small
\bibliographystyle{ieee_fullname}
\bibliography{references}
}

\end{document}